\documentclass[
]{ceurart}

\usepackage[T1]{fontenc}
\usepackage{graphicx}
\usepackage{hyperref}
\usepackage{color}

\usepackage{minted}
\setmintedinline{breaklines}

\usepackage{xspace}

\usepackage{amssymb,mathtools}

\usepackage{bm,bbm}

\usepackage{xcolor}

\usepackage{subcaption}
\usepackage{booktabs}
\usepackage{multirow}

\usepackage{tikz}
\usetikzlibrary{shapes}

\usepackage[capitalise]{cleveref}
\crefname{equation}{}{}

\usepackage{diagbox}
\usepackage{centernot}

\usepackage{siunitx}
\usepackage{bbding}

\newcommand{\ak}{A\"it-Kaci\xspace}
\newcommand{\bpl}{Bousi$\sim$Prolog\xspace}
\newcommand{\osf}{OSF\xspace}
\newcommand{\sparql}{SPARQL\xspace}
\newcommand{\github}{\url{https://github.com/fuzzyosf/fosf}\xspace}
\newcommand{\lib}{\texttt{fosf}\xspace}
\newcommand{\demourl}{\url{https://fosf-demo.pages.dev/}\xspace}
\newcommand{\cedar}{CEDAR\xspace}

\newcommand{\red}[1]{\textcolor{red}{#1}}
\newcommand{\green}[1]{\textcolor{mygreen}{#1}}

\makeatletter
\newcommand{\mymathcolor}[2]{%
  \begingroup
  \colorlet{out}{.}\color{#1}#2\@ifnextchar_{\do@mathcolorsub}{\endgroup}%
}
\newcommand{\do@mathcolorsub}[2]{%
  _{{\color{out}#2}}\@ifnextchar^{\do@mathcolorsup}{\endgroup}%
}
\newcommand{\do@mathcolorsup}[2]{%
  ^{\color{out}#2}\endgroup
}
\makeatother

\newcommand\sortcol[1]{%
  \ifcolors \textcolor{sortcol}{#1}%
  \else     {#1} \fi
}
\newcommand\strcol[1]{%
  \ifcolors \textcolor{gray}{#1}%
  \else     {#1} \fi
}
\newcommand\natcol[1]{%
  \ifcolors \textcolor{red!60!black}{#1}%
  \else     {#1} \fi
}
\newcommand\tagcol[1]{%
  \ifcolors \textcolor{tagcol}{#1}%
  \else     {#1} \fi
}
\newcommand\featcol[1]{%
  \ifcolors \textcolor{featcol}{#1}%
  \else     {#1} \fi
}

\newcommand{\person}{\s[person]}
\newcommand{\prof}{\s[professor]}

\newcommand{\lecturer}{\s[lecturer]}
\newcommand{\teacher}{\s[teacher]}
\newcommand{\researcher}{\s[researcher]}
\newcommand{\rcenter}{\s[research\_center]}
\newcommand{\inst}{\s[institution]}
\newcommand{\uni}{\s[university]}

\newcommand{\teachat}{\f[teach\_at]}
\newcommand{\researchat}{\f[research\_at]}

\DeclareFontFamily{U} {MnSymbolA}{}
\DeclareFontShape{U}{MnSymbolA}{m}{n}{
  <-6> MnSymbolA5
  <6-7> MnSymbolA6
  <7-8> MnSymbolA7
  <8-9> MnSymbolA8
  <9-10> MnSymbolA9
  <10-12> MnSymbolA10
  <12-> MnSymbolA12}{}
\DeclareFontShape{U}{MnSymbolA}{b}{n}{
  <-6> MnSymbolA-Bold5
  <6-7> MnSymbolA-Bold6
  <7-8> MnSymbolA-Bold7
  <8-9> MnSymbolA-Bold8
  <9-10> MnSymbolA-Bold9
  <10-12> MnSymbolA-Bold10
  <12-> MnSymbolA-Bold12}{}
\DeclareSymbolFont{MnSyA} {U} {MnSymbolA}{m}{n}

\DeclareFontFamily{U} {MnSymbolC}{}
\DeclareFontShape{U}{MnSymbolC}{m}{n}{
  <-6> MnSymbolC5
  <6-7> MnSymbolC6
  <7-8> MnSymbolC7
  <8-9> MnSymbolC8
  <9-10> MnSymbolC9
  <10-12> MnSymbolC10
  <12-> MnSymbolC12}{}
\DeclareFontShape{U}{MnSymbolC}{b}{n}{
  <-6> MnSymbolC-Bold5
  <6-7> MnSymbolC-Bold6
  <7-8> MnSymbolC-Bold7
  <8-9> MnSymbolC-Bold8
  <9-10> MnSymbolC-Bold9
  <10-12> MnSymbolC-Bold10
  <12-> MnSymbolC-Bold12}{}
\DeclareSymbolFont{MnSyC} {U} {MnSymbolC}{m}{n}

\DeclareFontFamily{U} {MnSymbolF}{}
\DeclareFontShape{U}{MnSymbolF}{m}{n}{
  <-6> MnSymbolF5
  <6-7> MnSymbolF6
  <7-8> MnSymbolF7
  <8-9> MnSymbolF8
  <9-10> MnSymbolF9
  <10-12> MnSymbolF10
  <12-> MnSymbolF12}{}
\DeclareFontShape{U}{MnSymbolF}{b}{n}{
  <-6> MnSymbolF-Bold5
  <6-7> MnSymbolF-Bold6
  <7-8> MnSymbolF-Bold7
  <8-9> MnSymbolF-Bold8
  <9-10> MnSymbolF-Bold9
  <10-12> MnSymbolF-Bold10
  <12-> MnSymbolF-Bold12}{}
\DeclareSymbolFont{MnSyF} {U} {MnSymbolF}{m}{n}

\newcommand{\bigcdot}{\boldsymbol{\cdot}}
\DeclareMathSymbol{\oast}{\mathbin}{MnSyC}{101}
\DeclareMathSymbol{\upsquigarrow}{\mathrel}{MnSyA}{161}
\DeclareMathSymbol{\downsquigarrow}{\mathrel}{MnSyA}{163}

\DeclareMathSymbol{\cap}{\mathbin}{MnSyC}{57}
\DeclareMathSymbol{\cup}{\mathbin}{MnSyC}{56}
\let\bigcap\relax
\DeclareMathSymbol{\bigcap}{\mathop}{MnSyF}{29}
\let\bigcup\relax
\DeclareMathSymbol{\bigcup}{\mathop}{MnSyF}{31}

\DeclareMathSymbol{\capdot}{\mathbin}{MnSyC}{61}
\DeclareMathSymbol{\cupdot}{\mathbin}{MnSyC}{60}
\DeclareMathSymbol{\bigcapdot}{\mathop}{MnSyF}{33}
\DeclareMathSymbol{\bigcupdot}{\mathop}{MnSyF}{35}

\DeclareMathSymbol{\curlywedge}{\mathbin}{MnSyC}{50}
\DeclareMathSymbol{\curlyvee}{\mathbin}{MnSyC}{51}
\DeclareMathSymbol{\bigcurlywedge}{\mathop}{MnSyF}{17}
\DeclareMathSymbol{\bigcurlyvee}{\mathop}{MnSyF}{19}

\DeclareMathSymbol{\curlywedgedot}{\mathbin}{MnSyC}{52}
\DeclareMathSymbol{\curlyveedot}{\mathbin}{MnSyC}{53}
\DeclareMathSymbol{\bigcurlywedgedot}{\mathop}{MnSyF}{21}
\DeclareMathSymbol{\bigcurlyveedot}{\mathop}{MnSyF}{23}

\makeatletter
\providecommand{\leftsquigarrow}{%
  \mathrel{\mathpalette\reflect@squig\relax}%
}
\newcommand{\reflect@squig}[2]{%
  \reflectbox{$\m@th#1\rightsquigarrow$}%
}
\makeatother

\input{preamble/fuzzy}
\DeclarePairedDelimiter\fceil
  {\lceil}
  {\rceil^{\mkern-3.5mu\bigcdot}}
\DeclarePairedDelimiter\ffloor
  {\lfloor}
  {\rfloor_{\mkern-3.5mu\bigcdot}}

\makeatletter
\newcommand{\defeq}{\mathrel{\aban@defeq}}
\newcommand{\aban@defeq}{%
  \vbox{\offinterlineskip\check@mathfonts
    \ialign{\hfil##\hfil\cr
      \fontsize{\ssf@size}{\z@}\normalfont def\cr
      \noalign{\kern1\p@}
      $\m@th=$\cr
      \noalign{\kern-.5\fontdimen22\textfont2}
    }%
  }%
}
\makeatother

\newcommand{\Iff}{\Leftrightarrow}

\makeatletter
\newcommand{\defiff}{\mathrel{\aban@defiff}}
\newcommand{\aban@defiff}{%
  \vbox{\offinterlineskip\check@mathfonts
    \ialign{\hfil##\hfil\cr
      \fontsize{\ssf@size}{\z@}\normalfont def\cr
      \noalign{\kern2.5\p@}
      $\m@th\Iff$\cr
      \noalign{\kern-.5\fontdimen22\textfont2}
    }%
  }%
}
\makeatother

\newcommand\restr[2]{{%
  \left.\kern-\nulldelimiterspace %
  #1 %
  \vphantom{\big|} %
  \right|_{#2} %
}}

\input{preamble/misc}

\newcommand{\isa}{\preceq}
\newcommand{\synisa}{\trianglelefteq}

\newcommand{\isop}{\mathop{\isa}}
\newcommand{\meet}{\curlywedge}

\NewDocumentCommand{\poset}{ O{\isa} O{X} }{\ensuremath{(#2, #1)}}

\newcommand{\eq}{\approx}
\NewDocumentCommand{\eqclass}{ O{\eq} O{x} }{[#2]_{#1}}

\NewDocumentCommand{\quot}{ O{\eq} O{X} }{\ensuremath{{#2}_{/{#1}}}}

\newif\ifcolors%

\renewcommand{\S}{\sortcol{\mathcal{S}}}
\newcommand{\F}{\featcol{\mathcal{F}}}
\newcommand{\V}{\tagcol{\mathcal{V}}}

\definecolor{mygreen}{rgb}{0.0, 0.65, 0.2}
\definecolor{tagcol}{rgb}{0.72, 0.1, 0.26}
\definecolor{sortcol}{rgb}{0.06, 0.22, 0.61}
\definecolor{featcol}{rgb}{0.0, 0.45, 0.0}
\definecolor{relcol}{rgb}{0.8, 0.52, 0.0}
\definecolor{stringcol}{rgb}{0.1, 0.1, 0.1}

\newcommand{\s}[1][s]{\sortcol{#1}}
\newcommand{\su}[1][s]{\s[{s'}]}
\newcommand{\sn}[1][s]{\sortcol{\texttt{#1}}}
\newcommand{\cc}[1][c]{\s[c]}

\renewcommand{\si}[1][1]{\s[{{s_{#1}}}]}

\newcommand{\str}[1][s]{\strcol{\text{\lstinline|#1|}}}
\newcommand{\nat}[1][s]{\natcol{\text{\lstinline|#1|}}}
\newcommand{\bots}{\s[{\bot}]}
\newcommand{\tops}{\s[{\top}]}

\newcommand{\f}[1][f]{\featcol{#1}}
\newcommand{\fn}[1][f]{\featcol{\texttt{#1}}}

\newcommand{\fj}[1][1]{\f[f_{#1}]}

\newcommand{\X}[1][X]{\tagcol{#1}\xspace}
\newcommand{\Xn}[1][X]{\tagcol{\texttt{#1}}\xspace}
\newcommand{\Y}[1][Y]{\X[#1]}
\newcommand{\Z}[1][Z]{\X[#1]}
\newcommand{\Xj}[1][1]{\tagcol{X_{#1}}}

\NewDocumentCommand{\xs}{O{X} O{s}}{%
    \ensuremath{\tagcol{#1}:\sortcol{#2}}
}

\NewDocumentCommand{\osfce}{O{\X} O{\X[X']}}{%
    \ensuremath{#1 \doteq #2}
}

\NewDocumentCommand{\osfcf}{O{\X} O{\f} O{\X[X']}}{%
    \ensuremath{#1.#2 \doteq #3}
}

\setmintedinline{breaklines}

\newcommand{\iltt}[1]{\text{\mintinline[fontsize=\normalsize]{turtle}{#1}}}

\newcommand{\ii}{\mathcal{I}}

\renewcommand{\gg}{\mathcal{G}}

\NewDocumentCommand{\denot}{O{t} O{\ii}}{[\![#1]\!]^{#2}}
\NewDocumentCommand{\denota}{O{t} O{v}}{[\![#1]\!]^{\ii,#2}}
\NewDocumentCommand{\denoti}{O{t} O{\ii} }{[\![#1]\!]^{#2,\alpha}}

\NewDocumentCommand{\tgraph}{O{f} O{g}}{G(\Z_{\f[#1],#2}:\tops)}
\NewDocumentCommand{\tvar}{O{f} O{g}}{\Z_{\f[#1],#2}}

\NewDocumentCommand{\fapproximates}{O{\gg}}{\mathop{\sqsubseteq^{#1}}}
\NewDocumentCommand{\fapprel}{O{\gg} O{\beta}}{\mathbin{\sqsubseteq^{#1}_{#2}}}

\colorstrue

\begin{document}

\copyrightyear{2026}
\copyrightclause{Copyright for this paper by its authors. Use permitted under Creative Commons License Attribution 4.0 International (CC BY 4.0).}
\conference{ISWC 2026 Companion Volume, October 25--29, 2026, Bari, Italy}

\title{Similarity-based OSF Logic for Flexible SPARQL Query Answering}

\author[1]{Gian Carlo Milanese}[%
orcid=0000-0002-1757-0152,
email=giancarlo.milanese@unimib.it
]
\cormark[1]

\author[1]{Gabriella Pasi}[%
orcid=0000-0002-6080-8170,
email=gabriella.pasi@unimib.it
]

\address[1]{University of Milano-Bicocca, DISCo, Milan, Italy}

\cortext[1]{Corresponding author.}

\begin{abstract}
  We present an updated version of \lib, a reasoner for fuzzy order-sorted
  feature (\osf) logic, with support for flexible \sparql query answering
  over RDF knowledge graphs.
  Fuzzy \osf logic is a knowledge representation and
  reasoning language where feature symbols denote functions and sort
  symbols denote fuzzy sets. Sorts are organized in a fuzzy subsumption
  relation that extends to \osf terms, record-like structures representing
  classes of entities. The unification algorithm for these structures
  provides a calculus of fuzzy type subsumption. The latest release of \lib
  introduces support for similarity-based reasoning by extending the \osf
  language with a similarity relation among sorts, alongside a translation engine
  for compiling \osf terms into standard SPARQL. Queries expressed as \osf terms
  undergo fuzzy ontology-aware rewriting prior to translation.
  This process incorporates an \osf theory of sort
  definitions, feature domain and range declarations, and fuzzy subsumption
  and similarity relations through fuzzy \osf constraint normalization.
  This rewriting enables flexible query answering by returning approximate
  solutions, allowing queries that would otherwise be inconsistent with the
  underlying ontology to yield relevant results.
\end{abstract}


\maketitle

\section{Introduction}%
\label{sec:intro}

We present an updated release of \lib,%
\footnote{Source code and examples: \github{}. Interactive demo: \demourl{}.}
a reasoner for fuzzy order-sorted
feature (\osf) logic, that introduces similarity-based \osf reasoning
alongside an ontology-aware translation engine for querying RDF knowledge
graphs (KGs) via \sparql.
Queries expressed as \osf terms undergo fuzzy constraint
normalization prior to translation into standard \sparql. This process
integrates ontological information from a sort taxonomy, a sort similarity
relation, sort definitions, and feature domain and range declarations. The
resulting query rewriting mechanism enables the retrieval of approximate
solutions for queries that would otherwise fail under strict ontological
constraints.

\paragraph{Background}%
\osf logic is a knowledge representation and reasoning language
introduced by \ak \cite{AitKaci1984,AitKaci1993b}.
It is based on \textit{sorts}, which represent classes of
objects,
and \textit{features},
which represent functional attributes. %
Sorts
are organized in a subsumption relation that extends to \osf terms,
record-like structures representing classes of entities and their
constraints.
Reasoning with \osf logic is based on the \textit{normalization} (or
\textit{unification}) of \osf terms, a procedure that aims to combine the
constraints expressed by one or more terms while taking
the sort hierarchy into account.
\osf logic is at the basis of CEDAR \cite{AmirAitKaci2017}, a
Semantic Web reasoner that normalizes and optimizes queries expressed as
\osf terms before translating them into SPARQL, possibly leading to more efficient
retrieval.

\textit{Fuzzy \osf logic} \cite{Milanese2024} extends \osf
logic by interpreting sorts and \osf terms as \textit{fuzzy
  sets}, and by generalizing the sort subsumption to a
\textit{fuzzy subsumption relation among sorts}, which is then extended to \osf terms.
The core reasoning algorithms of fuzzy \osf logic are implemented in the
\lib Python library \cite{Milanese2026}, which
supports fuzzy taxonomic reasoning, subsumption degree computation, and
fuzzy normalization and unification of \osf terms.

\paragraph{Contributions.}
The new release of \lib introduces the following features:
(i) support for a similarity relation between sorts, which is
integrated into the main reasoning algorithms of fuzzy \osf logic,
enabling \textit{similarity-based reasoning with \osf logic}
\cite{Milanese2024b};
(ii) support for feature domain and range declarations
during \osf term normalization;
(iii) querying RDF KGs through the translation of
normalized \osf queries into standard SPARQL.
The normalization exploits fuzzy sort subsumption and sort similarity
relations to relax incompatible constraints and retrieve approximate
solutions. %

\paragraph{Related work.}
Implementations of (crisp) \osf logic include the constraint logic
programming language LIFE \cite{AitKaci1993a} and the CEDAR Semantic Web
reasoner \cite{AmirAitKaci2017}. Closely related formalisms include
description logics (DLs) and fuzzy DLs, which are supported by
several reasoners (e.g.,
\cite{Tsarkov2006,Shearer2008,Sirin2007,Stoilos2006,BobilloStraccia2016}).
The similarities and differences between \osf logic and DLs have been
studied extensively \cite{Nebel1990,Nebel1991,AitKaci2007b,Krieger2010}.
Similarity-based \osf logic is inspired by fuzzy unification in fuzzy logic
programming, where unification is guided by similarity or proximity
relations (e.g., \cite{Gerla1999,Sessa2002,Iranzo2021,Kutsia2020,AitKaciPasi2020});
implementations include \bpl \cite{Iranzo2023} and FASILL
\cite{Iranzo2020fasill}.
In the Semantic Web community, various approaches handle imprecise or
flexible querying over RDF data, including fuzzy SPARQL extensions (e.g.,
\cite{Cheng2010,AlmendrosJimenez2024}) and query relaxation techniques
(e.g., \cite{Fakih2024}). In this context, \lib applies similarity-based
normalization at the \osf logic level before compiling the resulting query
terms into standard SPARQL.

\section{Similarity-based \osf Taxonomic Reasoning}%
\label{sec:simtax}

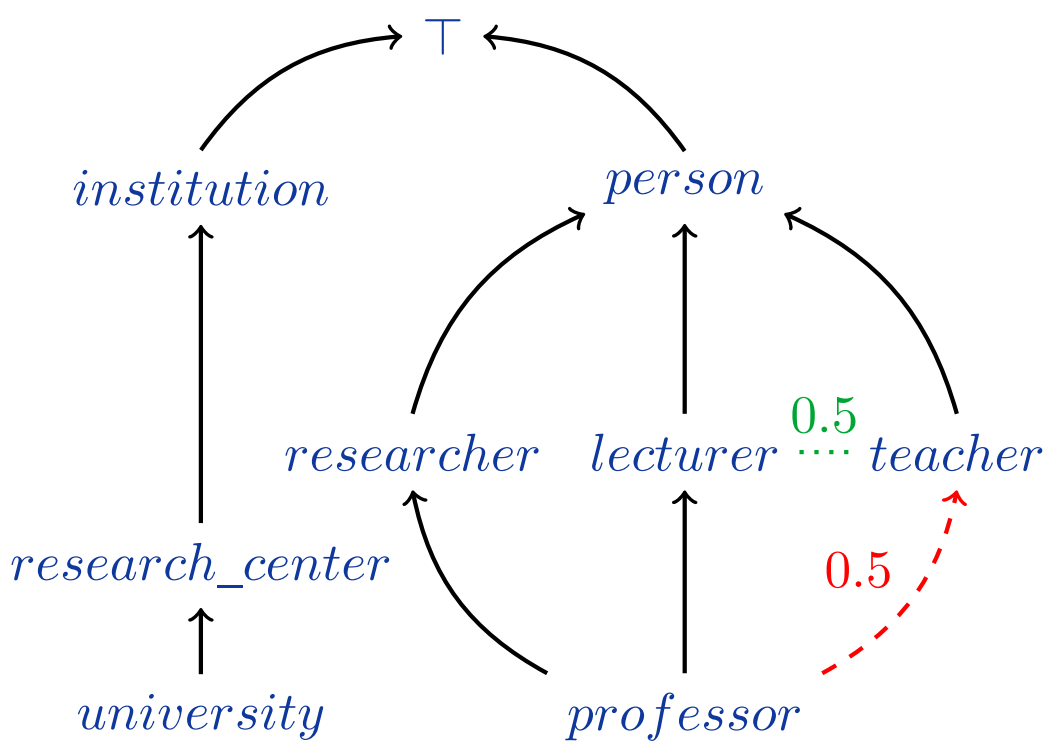
\begin{figure}[t!]
  \centering
  \small
  \begin{subfigure}{.48\textwidth}
    \begin{minted}{python}
from fosf.parsers import parse_taxonomy
sim_tax_str = """
# Subsumption declarations
institution, person < top .
research_center < institution .
university < research_center .
researcher, lecturer, teacher < person .
professor < researcher, lecturer .
# Similarity declarations
teacher ~ lecturer = 0.5 ."""
sim_tax = parse_taxonomy(sim_tax_str)
    \end{minted}
    \caption{Parsing subsumption and similarity relations.}
    \label{code:sim_tax}
  \end{subfigure}
  \begin{subfigure}{.48\textwidth}
    \begin{tikzpicture}[scale=1]
      \node[thick,draw=white] (top) at (2.4, 5) {$\tops$};

      \node[thick,draw=white] (uni) at (0.8, 0.5) {$\s[university]$};
      \node[thick,draw=white] (cen) at (0.8, 1.5) {$\s[research\_center]$}; %
      \node[thick,draw=white] (ins) at (0.8, 4.0) {$\s[institution]$};

      \node[thick,draw=white] (pro) at (4.0, 0.5) {$\s[professor]$};
      \node[thick,draw=white] (res) at (2.2, 2.25) {$\s[researcher]$};
      \node[thick,draw=white] (lec) at (4.0, 2.25) {$\s[lecturer]$};
      \node[thick,draw=white] (tea) at (5.8, 2.25) {$\s[teacher]$};
      \node[thick,draw=white] (per) at (4.0, 4.0) {$\s[person]$};

      \draw[->,thick] (uni.north) to node[left]   {} (cen.south);
      \draw[->,thick] (cen.north) to node[left]   {} (ins.south);
      \draw[-,thick,mygreen,dotted] (lec.east) to node[above] {$\green{0.5}$} (tea.west);
      \draw[->,thick,red,dashed] (pro.north east) to[bend right=25] node[left,yshift=1.9mm] {$\red{0.5}$} (tea.south);
      \draw[->,thick] (pro.north west) to[bend left=25] node[left] {} (res.south);

      \draw[->,thick] (pro.north) to node[right]   {} (lec.south);
      \draw[->,thick] (lec.north) to node[right]   {} (per.south);
      \draw[->,thick] (res.north) to[bend left=25] (per);
      \draw[->,thick] (tea.north) to[bend right=25] (per);
      \draw[->,thick] (ins.north) to[bend left=25] node[left]   {} (top.west);
      \draw[->,thick] (per.north) to[bend right=25] node[right]   {} (top.east);
    \end{tikzpicture}
    \label{fig:fuzzy_sub}
    \caption{Graphical representation.}
  \end{subfigure}
  \caption{Example of how a sort subsumption $\isop$ (black edges) and a sort similarity
    $\simop$ (dotted green edges) are parsed in \lib, and their graphical
    representation (where $\bots$ is omitted). After parsing, the two
    relations are combined into a single fuzzy subsumption $\fisop$ (black edges
    and dashed red edges).
  }
  \label{fig:sim_tax}
\end{figure}

At the core of \osf logic is a sort taxonomy, a finite
set of sort symbols $\S$ that is ordered in a lattice
$(\S, \isa, \meet)$
with top element
$\tops$ and bottom element
$\bots$. Similarity-based \osf logic \cite{Milanese2024b} extends the sort
taxonomy with a \textit{similarity relation} on sort symbols, that is, a
fuzzy binary relation $\sim: \S\times\S \to [0,1]$ which satisfies
reflexivity, symmetry and transitivity.\footnote{For fuzzy set theory
  definitions using the same notation as this paper, we refer to the appendix
  of \cite{Milanese2024b}.} As described in \cite{Milanese2024b}, the
information provided by the sort subsumption $\isa$ and the similarity
$\sim$ can be combined into a single fuzzy subsumption relation $\fisop :
\S\times\S \to [0,1]$, according to the following informal inference,
inspired by the similarity-based approaches to logic programming
(e.g., \cite{Sessa2002}):
\begin{center}
  \begin{tabular}{ll}
    if the sort $\si[0]$ is subsumed by the sort $\si[1]$          & (i.e., $\si[0]\isa\si[1]$)                  \\
    and $\si$ is similar to the sort $\si[2]$ with degree $\alpha$ & (i.e.,
    $\simop(\si, \si[2]) = \alpha$)                                                                                \\
    \hline
    then $\si[0]$ is subsumed by $\si[2]$ with degree $\alpha$     & (i.e., $\fisop(\si[0], \si[2]) = \alpha$).
  \end{tabular}
\end{center}
For example, if $\prof$s are $\lecturer$s, and $\lecturer$s are
similar to $\teacher$s with degree $0.5$, then we can conclude that
$\prof$s are also $\teacher$s with subsumption degree $0.5$.
As shown in \cite{Milanese2024b}, the resulting fuzzy subsumption relation
is a fuzzy lattice $(\S, \fisa, \fmeet)$, a fuzzy binary relation that satisfies
reflexivity, transitivity and strong antisymmetry, and such that
a GLB $\si[0]\fmeet\si[1]$ exists for every
$\si[0],\si[1]\in\S$.\footnote{A remark about notation: $\isa$ denotes the crisp subsumption
lattice on $\S$, and $\meet$ is the GLB operation on $(\S,
\isa)$; $\fisop$ is the fuzzy lattice resulting from combining $\isa$ with
$\sim$, and $\fmeet$ is the GLB operation on $(\S, \fisop)$.}

As shown in \cref{fig:sim_tax}, \lib supports parsing a sort
subsumption relation $\isop$ and a sort similarity relation $\sim$ and
combining them in a single fuzzy subsumption relation $\fisop$.
Cycle detection is performed to ensure that the resulting fuzzy
subsumption $\fisop$ is antisymmetric, possibly merging each sort cycle into
a single disjunctive set of sorts, according to \cite{Milanese2024b}.
When the resulting fuzzy relation $\fisop$ is not a (fuzzy)
lattice, but simply a partial order, \lib works on a completion of $\fisop$
that does not need to be explicitly computed.\footnote{%
  For more details, see \cite{CEDAR2,Milanese2024b}.
  In this paper we assume that $\fisop$ is always a fuzzy lattice for
  simplicity.
}
By employing the graph encoding
technique of \cite{AitKaci1989} and optimized weighted graph traversal of
\cite{Milanese2021a}, it is possible to efficiently compute the GLB
$\si[0]\fmeet\si[1]$ and the subsumption degree  $\fisop(\si[0], \si[1])$
of two sorts $\si[0]$ and $\si[1]$ \cite{Milanese2026}.
For instance,
$\teacher\fmeet\researcher = \prof$ and $\fisop(\prof, \teacher) = 0.5$.

\section{Similarity-based \osf Term Normalization}%
\label{sec:sim_unif}

\begin{figure}[t!]
  \begin{minted}[fontsize=\small]{python}
from fosf.parsers import parse_term
from fosf.reasoning import normalize_term
# Normalizing an OSF term according to subsumption + similarity only
t_str = "X0:person(teach_at -> X1:institution, research_at -> X1:research_center)"
t = parse_term(t_str)
print(normalize_term(t, sim_tax))
# Output: X0 : person(teach_at -> X1 : research_center, research_at -> X1)
  \end{minted}
  \caption{Normalizing an OSF term in \lib.}
  \label{code:norm}
\end{figure}

\begin{figure}[t!]
  \begin{minted}[fontsize=\small]{python}
from fosf.parsers import parse_theory
theory_str = sim_tax_str + """
professor           := Yp:professor(research_at -> Y1:university) .
domain(teach_at)    := teacher .
domain(research_at) := researcher ."""
theory = parse_theory(theory_str, ensure_closed=True)
# Normalizing an OSF term also according to the theory
nt, degree = normalize_term(t, sim_tax, theory, return_degree=True)
print(degree, nt)
# Output: 0.5 X0 : professor(teach_at -> X1 : university, research_at -> X1)
  \end{minted}
  \caption{Normalizing an \osf term modulo an \osf theory in \lib.}
  \label{code:theory}
\end{figure}

Sort symbols may be used with variables (also called tags)
$\Xj[0], \Xj[1] \ldots\in\V$ and features symbols $\fj[0], \ldots,
  \fj[n]\in\F$ to build
\textit{\osf terms}. These are record-like
representations of (fuzzy) sets of objects, such as
$t = \Xj[0]:\s[person]
  (
  \f[teach\_at] \to \Xj[1] : \s[institution],
  \f[research\_at] \to \Xj[1] : \s[research\_center]
  )$,
which denotes the class of people
that teach at some institution $\Xj[1]$ and carry out research at the same
$\Xj[1]$, which is also a research center.
In general, \osf terms may include
redundant, or even contradictory, information (e.g., the sort of
$\Xj[1]$ is specified twice in $t$).
\osf terms that are well-behaved to this regard
are called \textit{normal \osf terms} or \textit{$\psi$-terms}
\cite{AitKaci1993b}, and they are also denoted
$\psi$, $\psi_i$, and so on. The \lib reasoner supports normalizing \osf
terms according to a sort subsumption relation $\isop$ and similarity
$\sim$: as shown in \cref{code:norm}, the term $t$ can be normalized to
$\s[person]
  (
  \f[teach\_at] \to \Xj[1] : \s[research\_center],
  \f[research\_at] \to \Xj[1]
  )$, where the sort of $\Xj[1]$ is simplified to $\rcenter =
  \inst\fmeet\rcenter$.

The \lib reasoner also supports the normalization of \osf terms modulo an \textit{\osf
  theory} of \textit{sort definitions} and \textit{feature domain and range
  declarations}. Sort definitions associate each sort $\s$ with a normal term
$\Theta(\s) = \psi$
such that $\Theta(\bots) = \bots$, $\Theta(\tops) =
  \tops$, and the root sort of $\Theta(\s)$ is $\s$ for all $\s\in\S$
\cite{AitKaci1997}. Intuitively, the \textit{sort definition} $\Theta(\s)$
specifies structural constraints that each \osf term of type $\s$ should
satisfy. %
Feature \textit{domain} and \textit{range} declarations have shape
$\domain(\f) = \si[0]$ and $\range(\f) = \si[1]$, and they ensure that,
whenever $t' = \X:\tops(\f \to \Y:\tops)$ is an \osf term, then $\X$ must be of sort
$\si[0]$ and $\Y$ must be of sort $\si[1]$.
Overall, the (fuzzy) sort subsumption and similarity relations, the sort
definitions, and the feature domain and range declarations constitute the
\textit{ontological component} according to which an \osf term can be
normalized.
\cref{code:theory} shows an example of parsing an \osf theory in \lib
and normalizing the \osf term $t$ according to the theory.
In the example, the domain declarations for the features $\teachat$ and
$\researchat$ enforce the tag $\Xj[0]$ to be of sort $\teacher \fmeet
\researcher = \prof$, and the sort definition for $\prof$ thus
enforces $\Xj[1]$ to be a $\uni$.
The constraints imposed by an \osf theory may be
satisfied up to a degree $\alpha\in[0,1]$.
For instance, in the term \texttt{nt} of
\cref{code:theory}, $\Xj[0]$  is of sort $\prof$ rather than $\teacher$ (as
required by the domain of $\teachat$) so that \texttt{nt}
satisfies the \osf theory with degree $\fisop(\prof, \teacher) = 0.5$.
Note that, without the fuzzy subsumption $\fisop(\prof, \teacher)$ induced
by
$\prof \isa \lecturer$ and
$\simop(\lecturer, \teacher)$, the GLB of $\teacher$ and $\researcher$
would instead be $\bots$, and the normalization of $t$ would fail.

\section{Similarity-based OSF for SPARQL Query Answering}%
\label{sec:sim_sparql}

The \lib reasoner supports translating a (normalized) \osf term into \sparql
for instance retrieval over an RDF KG. This translation is inspired by the
query rewriting pipeline of \cedar
\cite{AmirAitKaci2017}, in which an \osf term query is preprocessed to
ensure consistency and to potentially reduce the retrieval search space
before generating a \sparql query. The \lib library extends this
framework by integrating sort definitions and the sort similarity
relation $\sim$ into the normalization, enabling
similarity-based query rewriting. Without the sort similarity, queries
violating ontological constraints would be declared inconsistent and
return no results. %
Instead, \lib can
rewrite \osf terms into queries that approximately satisfy the \osf theory
to some degree, thus enabling the retrieval of approximate solutions.

\begin{SaveVerbatim}[commandchars=\\\{\}]{Query}
SELECT \Xn[?X0] WHERE \{
  \Xn[?X0] rdf:type \sn[:person] .
  \Xn[?X0] \fn[:teach_at] \Xn[?X1] .
  \Xn[?X0] \fn[:research_at] \Xn[?X1] .
  \Xn[?X1] rdf:type \sn[:institution] .
  \Xn[?X1] rdf:type \sn[:research_center] . \}
\end{SaveVerbatim}

\begin{SaveVerbatim}[commandchars=\\\{\}]{OptimQuery}
SELECT \Xn[?X0] WHERE \{
  \Xn[?X0] rdf:type \sn[:professor] .
  \Xn[?X0] \fn[:teach_at] \Xn[?X1] .
  \Xn[?X0] \fn[:research_at] \Xn[?X1] .
  \Xn[?X1] rdf:type \sn[:university] . \}
\end{SaveVerbatim}

\begin{figure}[t!]
  \centering
  \small
  \begin{subfigure}{.48\textwidth}
    \centering
  \fbox{\BUseVerbatim{Query}}
  \caption{\sparql query before fuzzy \osf normalization.}
  \label{fig:unoptimsparql}
  \end{subfigure}
  \begin{subfigure}{.48\textwidth}
    \centering
  \fbox{\BUseVerbatim{OptimQuery}}
  \caption{\sparql query after fuzzy \osf normalization.}
  \label{fig:optimsparql}
  \end{subfigure}
  \caption{\sparql queries corresponding to the \osf term $t$, before and
  after normalization.}
  \label{fig:sparql}
\end{figure}

Consider again the query term $t = \X[?X_{0}]:\s[person] ( \f[teach\_at]
\to \Xj[1] : \s[institution], \f[research\_at] \to \Xj[1] :
\s[research\_center])$, where $\Xj[0]$ is marked with a question mark to
denote instance retrieval. Under \textit{standard \osf normalization}, $t$ is
inconsistent because $\teacher \meet \researcher = \bots$, as discussed in
\cref{sec:sim_unif}. Translating $t$ \textit{directly to \sparql
without normalizing} ignores the ontology entirely, yielding the query in
\cref{fig:unoptimsparql}, which fails to enforce domain/range semantics and
sort definitions. In contrast, by taking $\sim$ into account \lib
normalizes $t$ to $\Xj[0]:\prof( \f[teach\_at] \to \Xj[1] : \s[university],
\f[research\_at] \to \Xj[1])$, deriving an approximate solution that
satisfies the \osf theory with a degree of $0.5$ (\cref{code:theory}).
\cref{fig:optimsparql} shows the resulting \sparql translation. Providing
approximate solutions for otherwise inconsistent queries can be desirable
in many scenarios. For example, users often do not possess full knowledge
of extensive ontologies, making an inconsistent result potentially
surprising. Moreover, the inconsistency of $t$ arises from the closed-world
assumption of \osf logic, where $\teacher\meet\researcher=\bots$ forces
disjointness of the two sorts. Relaxing this requirement can be valuable in contexts where
this assumption does not hold, such as settings where domain knowledge can
evolve.

From a practical standpoint, \lib handles \osf query parsing, normalization,
and translation to SPARQL. Base IRIs and prefixes can be specified prior
to an \osf theory or query using Turtle-like syntax (e.g., \iltt{@base
<IRI> . @prefix : <IRI> .}). The syntax for \osf theories (including
subsumption and similarity declarations) and \osf terms supports base sorts
like $\person$ (resolved via the base IRI), as well as absolute IRIs (e.g.,
\iltt{<http://example.org/person>}) and prefixed names (e.g.,
\iltt{ex:person}). %
The query generated by \lib can %
be executed
over in-memory RDF graphs or remote SPARQL
endpoints using the \texttt{rdflib}
and \texttt{SPARQLWrapper}
libraries,%
\footnote{Available at \url{https://github.com/RDFLib/rdflib} and
\url{https://github.com/RDFLib/sparqlwrapper}.} or sent to any
standard SPARQL query engine.

\section{Demo Interface}%
\label{sec:demo}

\begin{figure}[t!]
  \centering
  \includegraphics[width=\textwidth]{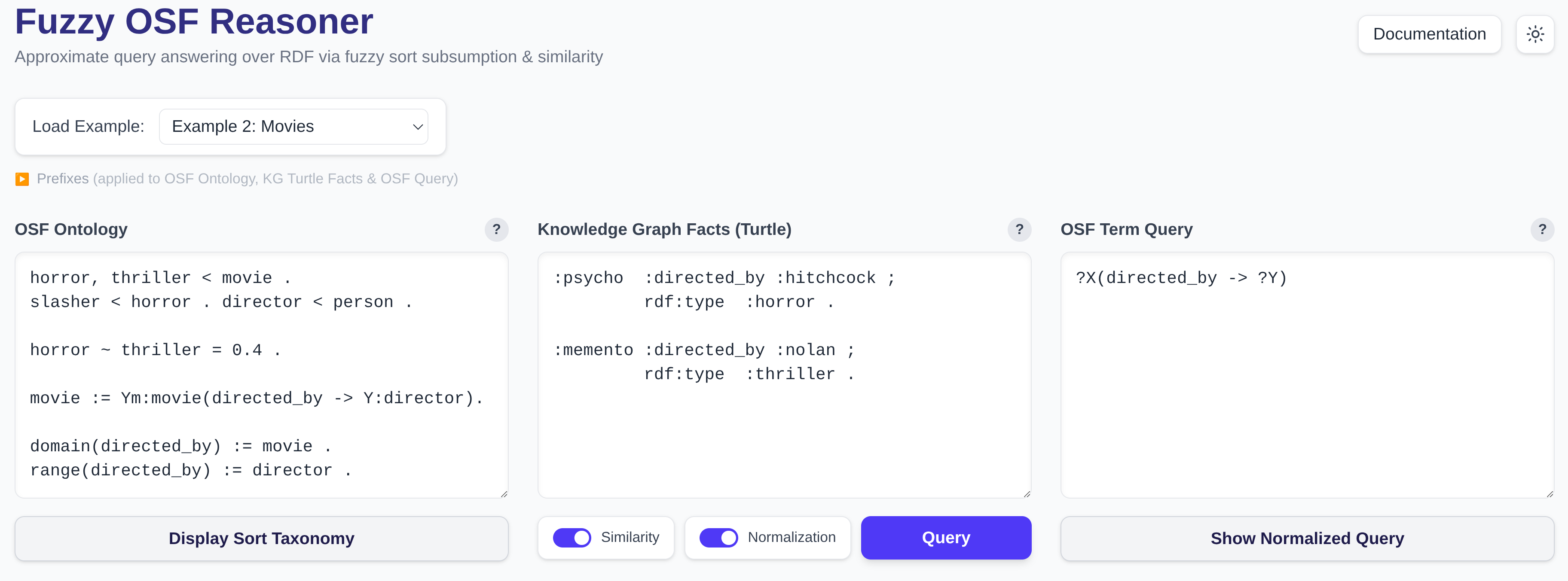}
\caption{The web interface of \lib, displaying the \osf ontology composed
  of sort subsumption and similarity declarations, sort definitions, and
  feature domain/range constraints (left), the target RDF KG facts (middle), and
  the \osf query term (right). Action buttons enable users to
inspect the taxonomy, display the translated query, and execute the
generated \sparql, with options to toggle \osf query normalization and the
similarity relation.}
  \label{fig:ui}
\end{figure}

We provide an interactive Web interface designed to demonstrate the query
normalization and execution pipeline of \lib.%
\footnote{Available at \demourl{}.}
As shown in \cref{fig:ui}, users can specify custom \osf theories, target
RDF KGs, and \osf queries directly in the browser, or select from a list of
pre-defined examples. The interface can visually render the sort
subsumption hierarchy, the sort similarity relation, and the resulting
fuzzy subsumption relation. Users can inspect the generated \sparql queries
and execute them directly against the RDF KG to view retrieved instances
alongside the approximation degree of the normalized query. Two global
toggles, for \osf query normalization and sort similarity, allow users to
compare \osf terms, generated \sparql queries, and retrieval results with or
without normalization and similarity-based reasoning.

\section{Conclusion and Future Work}%
\label{sec:conclusion}

We have presented an updated release of \lib, extending fuzzy \osf logic
with similarity-based reasoning and ontology-aware \sparql
query generation. By integrating a sort similarity relation into
constraint normalization, \lib enables flexible query rewriting, deriving
approximate solutions for queries that would otherwise be declared
inconsistent under strict ontological constraints.

While the current release of \lib provides a functional prototype, several
limitations present avenues for future development. Regarding
performance, we are replacing the pure-Python reasoning core with a C++
engine to improve execution speed and memory efficiency. To address the
closed-world assumption of \osf logic,
where two sorts $\si[0]$ and $\si[1]$ are treated as disjoint whenever
$\si[0]\fmeet\si[1]=\bots$,
we plan to allow declarations of non-disjoint
sorts within sort taxonomies, without the need to explicitly name their GLB.
Furthermore, we intend to improve the expressiveness of \lib by extending
support for disjunctive sorts and terms in both theories and queries, as
well as enabling sort instances in \osf query expressions. Finally, in
addition to manually defined similarity relations, we plan to support
automated similarity estimation using techniques such as vector embeddings.

\begin{acknowledgments}
  
The authors express their gratitude and a fond thought to Hassan \ak, who with
Gabriella Pasi set out to define a fuzzy version of \osf logic. Our development of
fuzzy \osf logic originates from their work on the definition of similarity-based
unification for OSF terms, extending the approach of \cite{AitKaciPasi2020}.
We also wish to thank Marco Braga and Stefano Molari for their help in
setting up the online demonstration system.

\end{acknowledgments}

\section*{Declaration on Generative AI}

During the preparation of this work, the author(s) used ChatGPT, Gemini
in order to: Grammar and spelling check. After using these tools, the
authors reviewed and edited the content as needed and take full
responsibility for the publication’s content.

\bibliography{main}

\end{document}